# Additive Tensor Decomposition Considering Structural Data Information

Shancong Mou, Andi Wang, Chuck Zhang and Jianjun Shi

*Abstract*—Tensor data with rich structural information becomes increasingly important in process modeling, monitoring, and diagnosis. Here structural information is referred to structural properties such as sparsity, smoothness, low-rank, and piecewise constancy. To reveal useful information from tensor data, we propose to decompose the tensor into the summation of multiple components based on different structural information of them. In this paper, we provide a new definition of structural information in tensor data. Based on it, we propose an additive tensor decomposition (ATD) framework to extract useful information from tensor data. This framework specifies a high dimensional optimization problem to obtain the components with distinct structural information. An alternating direction method of multipliers (ADMM) algorithm is proposed to solve it, which is highly parallelable and thus suitable for the proposed optimization problem. Two simulation examples and a real case study in medical image analysis illustrate the versatility and effectiveness of the ATD framework.

*Note to practitioners*—This paper was motivated by a real case in medical imaging which is the need of extracting aortic valve calcification (AVC) regions from the tensor data obtained from computed tomography (CT) image series of the aortic region. The main objective is to decompose image series into multiple components based on structural information. Similar needs are pervasive in medical image analysis as well as the image-based modeling, monitoring, and diagnosis of industrial processes and systems. Existing methods fail to incorporate a detailed description of the properties of the image series that reflect the physical understanding of the system in both the spatial and temporal domains. In this article, we provide a systematic description of the properties of image series and use them to develop a decomposition framework. It is applicable to various applications and can generate more accurate and more interpretable results.

## I. INTRODUCTION

W ITH the recent advancement of sensing technology, tensor data with rich structural information are acquired for process modeling, monitoring, and diagnosis [1]-[3]. For example, tensor data can represent multiple images obtained from CT scanning to illustrate different cross-sections of an organ [4]-[6]. In this example, the tensor is of order 3: the first mode represents multiple slices of images along the scanning direction, and the second and third modes define a cross-sectional image (Fig. 1).

Useful information for an application is usually deeply buried in tensor data. To reveal this information, we usually need to decompose the tensor data into a summation of multiple tensor components with the same size, based on domain interpretations. For example, the tensor representing the data shown in Fig. 1 can be decomposed into three tensor components, representing the background of blood and aorta, aortic valve calcification regions, and measurement errors. Generally, one of the tensor components represents quality issues of interest. In the example of the CT images, the component representing AVC regions is related to the likelihood of the paravalvular regurgitation (PVR) symptom after a surgery, which is a key quality index of transcatheter aortic valve replacement (TAVR), a common minimum invasive surgical procedure for treating aortic stenosis [6]. Based on the decomposed components, a prototype for patients' aortic root anatomies can be fabricated using multi-material 3D printing for surgical planning to prevent PVR. We refer to the decomposition of one tensor into a summation of multiple tensors as *additive tensor decomposition*, to differentiate it from the existing low-rank tensor decomposition approaches like Tucker and CP decomposition [7].

The decomposition of the tensor data can be performed based on different structural information of the components on various modes, driven by the physical knowledge underlying the process. In the CT example, we not only need to consider the spatial information in each image which are the second and third modes, such as every slice of the background tensor is smooth, the AVC regions are sparse, and measurement errors are usually small independent random values on each pixel. We also need to consider the temporal evolvement between images along the first tensor mode, such as gradual change of the component representing AVC regions across multiple slices. The structural difference between the components defines how the decomposition should be conducted. From the examples in Sections 3 and 4, we illustrate that the applications of ATD are

S. Mou is with the H. Milton Stewart School of Industrial and Systems Engineering, Georgia Institute of Technology, Atlanta, GA 30332 USA (e-mail: shancong.mou@gatech.edu).

A. Wang is with the H. Milton Stewart School of Industrial and Systems Engineering, Georgia Institute of Technology, Atlanta, GA 30332 USA (e-mail: andi.wang@gatech.edu).

C. Zhang is with the H. Milton Stewart School of Industrial and Systems Engineering and Georgia Tech Manufacturing Institute, Georgia Institute of Technology, Atlanta, GA 30332 USA (e-mail: chuck.zhang@gatech.edu).

J. Shi is with the H. Milton Stewart School of Industrial and Systems Engineering, Georgia Institute of Technology, Atlanta, GA 30332 USA (e-mail: jianjun.shi@isye.gatech.edu). Dr. Shi is the corresponding author.

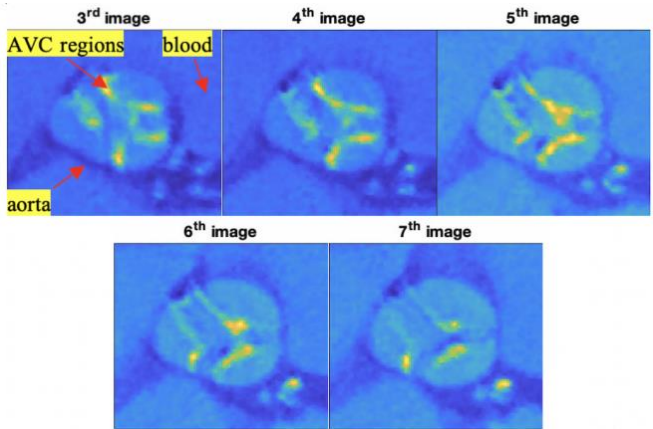

Fig. 1. A sample of sequential CT scans of the aortic root from the 3rd to the 7th image

prevalent in engineering, including the real-time monitoring of the crack growth on a building surface [8] and the overheating areas on industrial facilities [9].

In this study, we propose a systematic framework for additive tensor decomposition, which is based on a set of definitions of the structural information of the component tensors in various modes. As discussed in Section 3, this general framework utilizes the structural representations to specify a class of optimization problems that are easily customized for a wide range of tensor decomposition applications. In this framework, the optimization problems are defined as the summation of multiple terms that each term specifies one structural property of one tensor component. Such structural properties include the sparsity [10], smoothness [11], low-rank [12], and piecewise constancy which can be defined in one or multiple modes/slices of the tensor components. An alternating direction method of multipliers (ADMM) algorithm [13] is adopted for solving the problem, as it is highly parallelizable and thus suitable for high dimensional data analysis.

The rest of the paper is organized as follows. The related literature is reviewed in Section 2. Section 3 presents the definition of the structural information, the additive tensor decomposition framework, and the ADMM algorithm. The formulation and solution of two specific engineering examples are introduced to demonstrate the methodology. Section 4 further presents the simulation studies based on these two examples. In Section 5, a case study involving real data in medical imaging is presented, and Section 6 concludes the paper.

## II. LITERATURE REVIEW

Recently, some literature proposed the idea of decomposing a data matrix into a summation of multiple components. For identifying the anomalies in images, Yan et al. [1] proposed a smooth-sparse decomposition method to decompose an image into a smooth background and sparse anomalies. Similar additive decomposition is demonstrated in the Robust Principal Component Analysis (RPCA) [12], which decomposes the data matrix into a low-rank component and a sparse component.

The matrix decomposition framework has demonstrated its effectiveness in process monitoring [1, 2] and moving object detection [14]. As high dimensional tensors become increasingly common, there is an urgent need to extend this idea to tensor data. Compared with data matrices, a notable characteristic of tensor data is the distinct structural information of slices in certain modes. As a result, the tensor should be reshaped appropriately, so that the rich structural information can be revealed. Although some researchers extended the decomposition methods to tensor data, most of them did not incorporate this structural information into their decomposition strategies. For example, the Spatio-Temporal Smooth Sparse Decomposition (ST-SSD) method [2] extended the SSD [1] to handling Spatio-temporal tensor data. It only assumed that the decomposed two tensor components were either smooth or sparse in both time and space, but did not utilize these properties on individual modes. The RPCA was adjusted to identify the moving objects on a static background in image streams, while it was conducted by simply reshaping the spatio-temporal tensor to a matrix [14]. To extend the matrix decomposition framework to tensor data for generating more accurate and interpretable results, it is critical to have a systematic way to define the rich structural property of the tensor slices using mathematical formulations.

In the literature, assorted penalization methods have been used to promote the desired properties of the estimators in regularized regression. These penalties can be used in the tensor decomposition framework to enhance the smoothness in one or more slices of the tensor [11], promote the sparsity in different slices at one or more modes [10], and limit the patterns of variations in certain slices of the tensor [12]. Our proposed framework, ATD, combines these penalization methods with the unique structure of the tensor data into a high dimensional optimization problem, which can be solved efficiently with the ADMM algorithm [13].

## III. ADDITIVE TENSOR DECOMPOSITION FRAMEWORK

In this section, we introduce the general ATD framework and the solution procedure, and further demonstrate them using two examples. Throughout the article, the set $\{1, ..., n\}$ is represented as $[n]$. We denote a scalar by a lowercase letter $a$, a vector by a boldface lowercase letter $\mathbf{a}$, and a matrix by a boldface uppercase letter $\mathbf{A}$. An order-$n$ tensor is denoted by a calligraphic letter $\mathcal{A} \in \mathbb{R}^{I_1 \times I_2 \times \cdots \times I_d}$, where $I_i$ is the dimension of its $i$th mode. Element $(i_1, ..., i_d)$ of the tensor $\mathcal{A}$ is represented as $\mathcal{A}(i_1, ..., i_d)$. The $(i_{d_1}, ..., i_{d_k})$ slice of the tensor $\mathcal{A}$ at mode $d_1, ..., d_k$ is represented as $\mathcal{A}(:, ..., :, i_{d_1}, :, ..., :, i_{d_k}, :, ..., :)$, where $i_{d_j} \in [I_{d_j}]$ represents the elements at mode $d_j$, $j = 1, ..., k$. For ease of exposition, let $\mathcal{A}_{i_{d_1}, ..., i_{d_k}}$ denote $\mathcal{A}(:, ..., :, i_{d_1}, :, ..., :, i_{d_k}, :, ..., :)$ The mode-$k$ fiber is defined by a column vector $\mathcal{A}(i_1, ..., :, i_{k-1}, :, i_{k+1}, ..., i_d)$ [7]. The mode-$(r_1, ..., r_L)$ matricization of $\mathcal{A}$ is a matrix $\mathcal{A}_{(I_{r_1}\cdots I_{r_L})\times(I_{q_1}\cdots I_{q_{d-L}})} \in \mathbb{R}^{(I_{r_1}\cdots I_{r_L})\times(I_{q_1}\cdots I_{q_{d-L}})}$, whose element at entry $(i_{r_1}+(i_{r_2}-1)I_{r_1}+\cdots+(i_{r_L}-1)I_{r_1}\cdots I_{r_{L-1}}, i_{q_1}+(i_{q_2}-1)I_{q_1}+\cdots+(i_{q_{d-L}}-1)I_{q_1}\cdots I_{q_{d-L-1}})$ is $\mathcal{A}(i_1,\ldots,i_d)$, where $\{r_1,\ldots,r_L\}$ and $\{q_1,\ldots,q_{d-L}\}$ is a partition of $\{1,\ldots,d\}$ and $i_k \in [I_k]$ for all $k \in [d]$. Specifically, $\mathcal{A}_{(j)} \in \mathbb{R}^{I_j \times I_{-j}}$ is the mode-$j$ matricization of $\mathcal{A}$ whose columns are the slices of $\mathcal{A}$ at mode $j$, where $I_{-j} = \Pi_{l\in[d]-\{j\}}I_l$.

### A. General problem formulation

Assume that an order-$d$ tensor $\mathcal{M} \in \mathbb{R}^{I_1\times\cdots\times I_d}$ is decomposed into a summation of $m$ tensor components $\mathcal{X}_i$ with the same size, where $i \in [m]$, and that there are $n_i$ structural assumptions on $\mathcal{X}_i$. The decomposition is achieved by minimizing the summation of regularization terms specified by all structural properties given in Problem (1):

$$\underset{\mathcal{X}_1,\ldots,\mathcal{X}_m}{\text{minimize}} \sum_{i=1}^{m}\sum_{j=1}^{n_i}\lambda_{i,j}p_{i,j}(\mathcal{X}_i),$$
$$\text{subject to } \mathcal{M} = \sum_{i=1}^{m}\mathcal{X}_i. \qquad (1)$$

In Problem (1), $p_{i,j}(\cdot)$ specifies the $j$th structural property on $\mathcal{X}_i$, by taking a large value when $\mathcal{X}_i$ does not satisfy the specified structural properties. The trade-off between multiple regularization terms is specified by the tuning parameters $\lambda_{i,j}$'s. Certain structural properties are detailed below, which can be applied to either the entire tensor or certain slices.

**Structural property on smoothness.** The elements in a tensor component $\mathcal{X}$ or one of its slices should have similar values, whenever their indices are close to each other. In other words, the entire tensor $\mathcal{X}$ or the slices at certain modes should be smooth [1, 2]. We use mode-$(d_1,\ldots,d_k)$ smoothness of a tensor to describe the similarity among different slices at mode $d_1,\ldots,d_k$, which can be represented as $p(\mathcal{X}) = \sum_{s=1}^{k}\left\|\mathbf{D}_{d_s}^{l}\mathcal{X}_{(d_s)}\right\|_F^2$, where $\mathbf{D}_{d_s}^{l} \in \mathbb{R}^{I_{d_s}\times I_{d_s}}, s\in[k]$ are finite difference matrices of order $l$ on mode $s$ [1]. The notation $\|\mathbf{A}\|_F$ represents the Frobenius norm of $\mathbf{A}$. For example, when $\mathcal{X}$ is an order-3 tensor whose first mode represents time and every $\mathcal{X}(i_1,:,:)$ represents an image, the mode-(2,3) smoothness essentially indicates that each image is smooth (Fig. 2 (a)). The mode-(1) smoothness, on the other hand, indicates that all vectors $\mathcal{X}(:,i_2,i_3)$ compose a smooth curve for every element $(i_2,i_3)$, where $i_2 \in [I_2]$ and $i_3 \in [I_3]$. It means that the slices of $\mathcal{X}$ change smoothly in temporal mode (Fig. 2 (b)).

**Structural property on sparsity.** Some tensor components represent the anomaly and isolated features, and sparsity is needed for them. To reflect the understandings of the problem, the sparsity in a tensor may be specified mode-$(d_1,\ldots,d_k)$ sparsity. We use mode-$(d_1,\ldots,d_k)$ sparsity to describe the sparsity among slices at mode $d_1,\ldots,d_k$, which can be represented as $p(\mathcal{X}) = \sum_{i_{d_1}=1}^{I_{d_1}}\cdots\sum_{i_{d_k}=1}^{I_{d_k}}\left\|\text{vec}\left(\mathcal{X}(:,\ldots,:,i_{d_1},:,\ldots,:,i_{d_k},:,\ldots,:)\right)\right\|_2$, which is the group Lasso penalty. For example, the mode-(1) sparsity indicates that only some slices at mode 1 contain non-zero values (Fig. 3 (a)). The mode-(2,3) sparsity indicates that

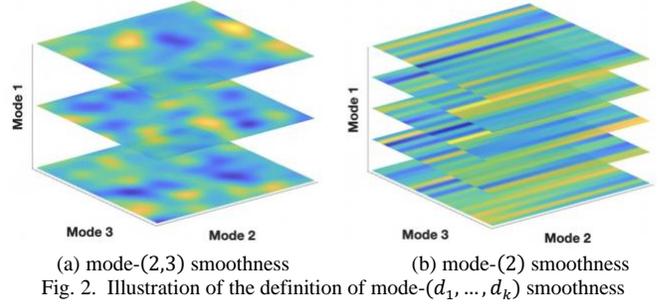

(a) mode-(2,3) smoothness  (b) mode-(2) smoothness
Fig. 2. Illustration of the definition of mode-$(d_1,\ldots,d_k)$ smoothness

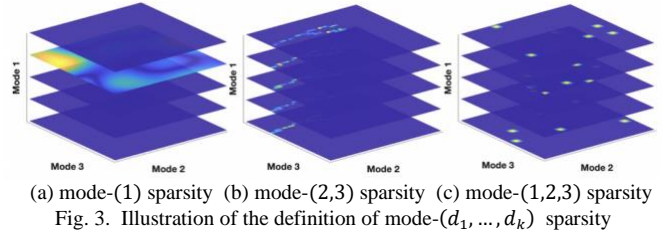

(a) mode-(1) sparsity  (b) mode-(2,3) sparsity  (c) mode-(1,2,3) sparsity
Fig. 3. Illustration of the definition of mode-$(d_1,\ldots,d_k)$ sparsity

each slice at mode 1 is sparse (Fig. 3 (b)). Note that mode-$(1,\ldots,d)$ sparsity is equivalent to the sparsity of the whole tensor as $p(\mathcal{X}) = \sum_{i_1=1}^{I_1}\cdots\sum_{i_d=1}^{I_d}\|\mathcal{X}(i_1,\ldots,i_d)\|_2 = \|\text{vec}(\mathcal{X})\|_1$, as indicated in Fig. 3 (c).

**Structural property on the variation patterns.** In many applications, certain modes of the tensor data have a limited number of variation patterns. In other words, appropriate reshaping operations should be applied to transform the associated slices of the tensor to low-rank matrices. We define the mode-$(l_1,\ldots,l_q)$ low rank of slices at mode $d_1,\ldots,d_k$ where $(l_1,\ldots,l_q) \cap (d_1,\ldots,d_k) = \emptyset$ as follows: let $\{s_1,\ldots,s_{d-q-k}\} = [d] - \{l_1,\ldots,l_q\} - \{d_1,\ldots,d_k\}$, the mode-$(l_1,\ldots,l_q)$ matricization of each slice at mode $d_1,\ldots,d_k$, $\mathcal{X}(:,\ldots,:,i_{d_1},:,\ldots,:,i_{d_k},:,\ldots,:)$ is low rank. Therefore, mode-$(l_1,\ldots,l_q)$ low rank of slices at mode $d_1,\ldots,d_k$ can be represented as

$$p(\mathcal{X}) = \sum_{i_{d_1}=1}^{I_{d_1}}\cdots\sum_{i_{d_k}=1}^{I_{d_k}}$$
$$\left\|\left(\mathcal{X}(:,\ldots,:,i_{d_1},:,\ldots,:,i_{d_k},:,\ldots,:)\right)_{(I_{l_1}\cdots I_{l_q})\times(I_{s_1}\cdots I_{s_{d-q-k}})}\right\|_*,$$

where $\|\cdot\|_*$ is the nuclear norm. We say that a tensor is mode-$(l_1,\ldots,l_q)$ low rank if its mode-$(l_1,\ldots,l_q)$ matricization is low rank. There are two examples for order-3 tensors whose first mode is time and the rest represent images. First, consider that each image in $\mathcal{X}$ represents a textured background with repetitive vertical and horizontal patterns [15]. Then each slice $\mathcal{X}(i_1,:,:)$ at mode 1 is of mode-(3) low rank (Fig. 4), and the tensor can be regularized by $p(\mathcal{X}) = \sum_{i_1=1}^{I_1}\|(\mathcal{X}(i_1,:,:))\|_*$, where $\|\cdot\|_*$ is the nuclear norm. In another example, all images in the tensor $\mathcal{X}$ are similar, representing a static background. Then $\mathcal{X}$ is of mode-(1) low rank and the regularization $p(\mathcal{X}) = \|\mathcal{X}_{(1)}\|_*$ should be employed.

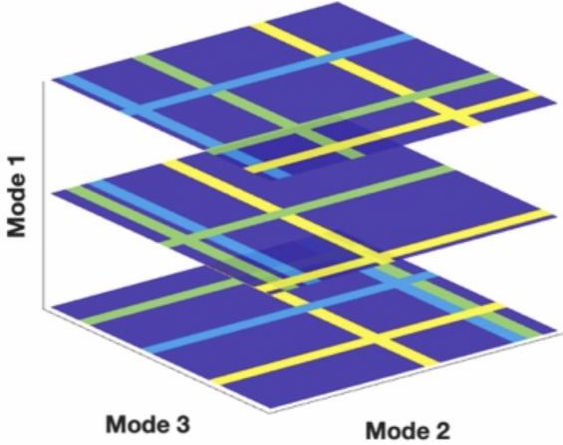

Fig. 4. mode-(3) low rank of slices at mode 1

**Structural information on piecewise constancy.** In certain applications, some slices of the tensor are piecewise constant. Without loss of generality, we formally define mode$-(1, \dots, q)$ piecewise constancy of slices at mode $q+1, \dots, q+k$ as follows:

*Definition:* $\mathcal{X}$ is mode-$(1, \dots, q)$ piecewise constancy of slices at mode $q+1, \dots, q+k$, if for any $i_{q+1} \in [I_{q+1}], \dots, i_{q+k} \in [I_{q+k}]$, the set of grid points $I_{q+k+1} \times \cdots \times I_{q+d}$ can be partitioned into multiple continuous regions $R_1(i_{q+1}, \dots, i_{q+k}), \dots, R_k(i_{q+1}, \dots, i_{q+k})$, such that any two order $q$ sub-tensors

$$\mathcal{X}(:, \dots, :, i_{q+1}, :, \dots, :, i_{q+k}, i_{q+k+1}, \dots, i_d)$$
$$= \mathcal{X}(:, \dots, :, i_{q+1}, :, \dots, :, i_{q+k}, i'_{q+k+1}, \dots, i'_d)$$

where $(i_{q+k+1}, \dots, i_d)$ and $(i'_{q+k+1}, \dots, i'_d)$ are in the one of the same regions.

This definition can be extended to mode-$(l_1, \dots, l_q)$ piecewise constancy of slices at mode $d_1, \dots, d_k$ naturally for arbitrary slices $(l_1, \dots, l_q) \cap (d_1, \dots, d_k) = \emptyset$. Then, mode-$(1, \dots, q)$ piecewise constancy of slices at mode $q+1, \dots, q+k$ can be represented as

$$p(\mathcal{X}) = \sum_{i_{q+1}=1}^{I_{q+1}} \cdots \sum_{i_{q+j}=1}^{I_{q+j}} \sum_{\left((i_{q+k+1}, \dots, i_d),(i'_{q+k+1}, \dots, i'_d)\right) \in E}$$
$$\left\| \text{vec}\left( \mathcal{X}(:, \dots, :, i_{q+1}, :, \dots, :, i_{q+k}, i_{q+k+1}, \dots, i_d) - \right.\right.$$
$$\left.\left. \mathcal{X}(:, \dots, :, i_{q+1}, :, \dots, :, i_{q+k}, i'_{q+k+1}, \dots, i'_d) \right) \right\|_2,$$

where $\left((i_{q+k+1}, \dots, i_d),(i'_{q+k+1}, \dots, i'_d)\right) \in E$ means that $(i_{q+k+1}, \dots, i_d)$ is in the neighborhood of $(i'_{q+k+1}, \dots, i'_d)$. Specifically, the neighborhood can be defined as

Queen-type neighborhood:
$E = \{((i_1, \dots, i_d),(i'_1, \dots, i'_d)): |i'_s - i_s| \le 1, s = 1, \dots, d\};$

Rook-type neighborhood:
$E = \{((i_1, \dots, i_d),(i'_1, \dots, i'_d)): |i'_s - i_s| = 1, 1 \le s \le d, i_j = i'_j \text{ for all } j \ne s\};$

In multiple change-point detections, most adjacent slices at temporal mode $d_s$ in slices at mode $(d_1, \dots, d_k)$ shall be the same if instantaneous change seldom happens, where $s \notin [k]$. It is defined as mode-$(d_s)$ piecewise constancy of slices at mode $d_1, \dots, d_k$. This characteristic of the tensor can be regularized using a fused lasso penalty [16],

$$p(\mathcal{X})$$
$$= \sum_{i_{d_1}=1}^{I_{d_1}} \cdots \sum_{i_{d_k}=1}^{I_{d_k}} \sum_{i_s=1}^{I_s-1} \left\| \text{vec}\left( \mathcal{X}(:, \dots, :, i_{d_1}, :, \dots, :, i_{d_k}, :, \dots, :, i_{d_s} \right.\right.$$
$$\left.\left. + 1, \dots, :) - \mathcal{X}(:, \dots, :, i_{d_1}, :, \dots, :, i_{d_k}, :, \dots, :, i_{d_s}, \dots, :)) \right) \right\|_2.$$

For example, when $\mathcal{X}$ is an order-3 tensor whose first mode represents time and every $\mathcal{X}(i_1, :, :)$ represents an image, mode-(1) piecewise constancy of $\mathcal{X}$ indicates that instantaneous change among images seldom happens at temporal mode (Fig. 5 (a)). For another example, mode-(2) piecewise constancy of slices at mode 1 means that it is piecewise constant inside each image along mode 2 (Fig. 5 (b)).

With the regularization terms promoting the above-mentioned properties, formulation (1) is versatile and can be tailored for many applications. Two specific examples are below.

*1) Example 1: Monitoring crack growth on the surface of engineering structures*

Engineering structures are often subject to fatigue stress which leads to crack in the structure materials such as concrete surfaces and beams. Image-based crack detection becomes popular due to its high efficiency and objective assessment of deterioration. However, the irregular size of cracks and irregularly illuminated conditions in the acquired images are the main challenges in this inspection method [17]. In this example, we show the capability of ATD method in monitoring the crack growth under irregular illuminated conditions. We take images of a concrete wall at a fixed orientation every day to monitor the growth of a crack on it. All images collected in $I_1$ days can be represented in a tensor $\mathcal{M} \in \mathbb{R}^{I_1 \times I_2 \times I_3}$, where $I_2 \times I_3$ is the size of each image. Our objective is to decompose $\mathcal{M}$ into a summation of two components, the background of the wall $\mathcal{X}_1$ and the crack $\mathcal{X}_2$. The backgrounds of the images are all smooth, but they are subject to the variation caused by the unstable brightness conditions. To describe the smoothness of each image, the regularization is represented as mode-(2,3) smoothness $p_{1,1}(\mathcal{X}_1) = \left\| \mathbf{D}_2^1 \mathcal{X}_{1(2)} \right\|_F^2$ and $p_{1,2}(\mathcal{X}_1) = \left\| \mathbf{D}_3^1 \mathcal{X}_{1(3)} \right\|_F^2$, where $\mathbf{D}_i^1 s \in \mathbb{R}^{l_i \times l_i}$ are first-order difference matrix [1] for $i = 1, \dots, 3$

$$\mathbf{D}_i^1 = \begin{bmatrix} 1 & -1 & & & \mathbf{0} \\ 1 & -1 & & & \\ & 1 & -1 & & \\ & & \ddots & \ddots & \\ & & & 1 & -1 \\ \mathbf{0} & & & & 1 & -1 \end{bmatrix}.$$

The crack is a consecutive line limited to a local region on the wall [8], and it grows slowly. It is subject to the variation caused by the unstable brightness conditions, therefore the pixel intensity on the same line can be different. Therefore, the temporal smoothness is represented as $\left\|\mathbf{D}_1^2 \mathcal{X}_{2(1)}\right\|_F^2$ which is mode-(1) smoothness, where $\mathbf{D}_1^2 \in \mathbb{R}^{I_1 \times I_1}$ is the second-order difference matrix with modified Neumann boundary condition [18],

$$\mathbf{D}_1^2 = \begin{bmatrix} -1 & 1 & & & & \mathbf{0} \\ 1 & -2 & 1 & & & \\ & 1 & -2 & 1 & & \\ & & \ddots & \ddots & \ddots & \\ & & & 1 & -2 & 1 \\ \mathbf{0} & & & & 1 & -1 \end{bmatrix}.$$

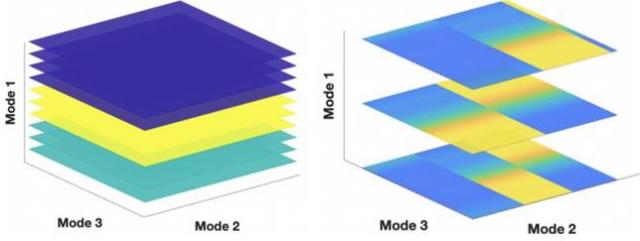

(a) mode-(1) piecewise constancy of $\mathcal{X}$    (b) mode-(2) piecewise constancy of slices at mode 1

Fig. 5. Illustration of the definition of mode-$(1, \ldots, q)$ piecewise constancy

The mode-(1,2,3) sparsity of the anomaly is represented as $\|\mathrm{vec}(\mathcal{X}_2)\|_1$. In total, the optimization problem specified by the ATD framework is given as

$$\begin{aligned} \underset{\mathcal{X}_1, \mathcal{X}_2}{\text{minimize}} \quad & \lambda_{1,1} \left\|\mathbf{D}_2^2 \mathcal{X}_{2(2)}\right\|_F^2 + \lambda_{1,2} \left\|\mathbf{D}_3^2 \mathcal{X}_{3(3)}\right\|_F^2 \\ & + \lambda_{2,1} \left\|\mathbf{D}_1^2 \mathcal{X}_{2(1)}\right\|_F^2 + \lambda_{2,2} \|\mathrm{vec}(\mathcal{X}_2)\|_1, \\ \text{subject to} \quad & \mathcal{M} = \mathcal{X}_1 + \mathcal{X}_2. \end{aligned} \tag{2}$$

where $\lambda_{1,1}, \lambda_{1,2}, \lambda_{2,1}, \lambda_{2,2}$ are tuning parameters.

### 2) Example 2: Monitoring the spots of overheating on a heated surface

High temperature is one of the root causes for equipment degradation. Infrared thermography (IRT) can provide the thermal image of the entire measured equipment. Therefore, it has been successfully utilized in condition monitoring in numerous industries including nuclear, aerospace, and paper industries [19]. In this example, we show the capability of ATD framework of monitoring the spot of overheating on a heated surface. A thermal camera was installed above a surface of a fluidized catalytic cracking regenerator, to monitor the spots of overheating [9]. Spots of overheating were observed as few small regions on the monitored surface in which the temperature is much higher than other sections of the surface. In general, some hotspots' locations may change in different time frames and others may appear at the same location throughout the monitoring period. Apart from the hotspots, the background temperature of the surface was smooth and varying slowly over time, driven by the heated material inside the container and the environmental condition that affects the cooling.

The collected thermal images form a tensor $\mathcal{M} \in \mathbb{R}^{I_1 \times I_2 \times I_3}$. Our objective is to decompose it into three tensors $\mathcal{X}_1$, $\mathcal{X}_2$, and $\mathcal{X}_3$ that represent the varying background, the static hotspots, and the moving hotspots respectively. Among them, all images in $\mathcal{X}_1$ are smooth, and this property is regularized by mode-

(2,3) smoothness $p_{1,1}(\mathcal{X}_1) = \left\|\mathbf{D}_2^2 \mathcal{X}_{1(2)}\right\|_F^2$ and $p_{1,2}(\mathcal{X}_1) = \left\|\mathbf{D}_3^2 \mathcal{X}_{1(3)}\right\|_F^2$. Also, variations of the images of the background are only driven by the heating and cooling effects, and thus all background images reside in a low-rank subspace. Therefore, $\mathcal{X}_{1(1)}$ should have the low-rank property and another regularization $p_{1,3}(\mathcal{X}_1) = \|\mathcal{X}_{1(1)}\|_*$ is employed. Most elements in $\mathcal{X}_2$ are zero, which is regularized by $p_{2,2}(\mathcal{X}_2) = \|\mathrm{vec}(\mathcal{X}_2)\|_1$. Also, the images in $\mathcal{X}_2$ are similar, which means that they should also reside in a low-dimensional subspace, and thus $\mathcal{X}_2$ is also regularized by mode-(1) low rank penalty $p_{2,1}(\mathcal{X}_2) = \|\mathcal{X}_{2(1)}\|_*$. Finally, most elements in $\mathcal{X}_3$ are zero, so the regularization on $\mathcal{X}_3$ is given by $p_{3,1}(\mathcal{X}_3) = \|\mathrm{vec}(\mathcal{X}_3)\|_1$. Put everything together, we formulate Problem (3) using the ATD framework

$$\begin{aligned} \underset{\mathcal{X}_1, \mathcal{X}_2, \mathcal{X}_3}{\text{minimize}} \quad & \lambda_{1,1} \left\|\mathbf{D}_2^2 \mathcal{X}_{1(2)}\right\|_F^2 + \lambda_{1,2} \left\|\mathbf{D}_3^2 \mathcal{X}_{1(3)}\right\|_F^2 + \lambda_{1,3} \|\mathcal{X}_{1(1)}\|_* \\ & + \lambda_{2,1} \|\mathcal{X}_{2(1)}\|_* + \lambda_{2,2} \|\mathrm{vec}(\mathcal{X}_2)\|_1 + \lambda_{3,1} \|\mathrm{vec}(\mathcal{X}_3)\|_1, \\ \text{subject to} \quad & \mathcal{M} = \mathcal{X}_1 + \mathcal{X}_2 + \mathcal{X}_3, \end{aligned} \tag{3}$$

where $\lambda_{1,1}, \lambda_{1,2}, \lambda_{1,3}, \lambda_{2,1}, \lambda_{2,2}$ and $\lambda_{3,1}$ are tuning parameters.

We will follow up on solution procedures, simulated images (Fig. 6 and Fig. 7), and simulation results for the Problems (2) and (3) in the latter part of the paper.

### B. Problem solution

Notice that Problem (1) is convex and bounded from below by zero. Therefore, an optimal solution to this problem exists. To deal with the high dimensionality of the decision variables, we adopt an ADMM algorithm to solve this problem [13]. In Problem (1), there are $n_i$ additive terms associated with the same variable $\mathcal{X}_i$. We introduce $n_i$ new ancillary tensors $\mathcal{X}_i^{(1)}, \ldots, \mathcal{X}_i^{(n_i)}$ as copies of $\mathcal{X}_i$, and further let $\widetilde{\mathcal{X}}_i = \left(\mathcal{X}_i^{(1)}, \ldots, \mathcal{X}_i^{(n_i)}\right)$ and $\widetilde{\mathcal{X}} = (\widetilde{\mathcal{X}}_1, \ldots, \widetilde{\mathcal{X}}_m)$. Then, formulation (1) is transformed into

$$\text{minimize} \quad f(\widetilde{\mathcal{X}}) + g(\widetilde{\mathcal{X}}), \tag{4}$$

where $f(\widetilde{\mathcal{X}}) = \sum_{i=1}^m \sum_{j=1}^{n_i} \lambda_{i,j} p_{i,j}\left(\mathcal{X}_i^{(j)}\right)$, and $g(\widetilde{\mathcal{X}}) = I_C(\widetilde{\mathcal{X}})$. Here $I_\Omega(x)$ refers to an indicator function that takes value 0 when $x \in \Omega$, and takes value $+\infty$ if $x \notin \Omega$. The set $C = \left\{\mathcal{X}_i^{(1)} = \cdots = \mathcal{X}_i^{(n_i)}, i \in [m]\right\} \cap \left\{\mathcal{M} = \sum_{i=1}^m \mathcal{X}_i^{(1)}\right\}$. The canonical form (4) can be solved using the ADMM algorithm listed in Algorithm 1.

---

**Algorithm 1** ADMM algorithm

Initialize $\widetilde{\mathcal{Z}}$ and $\widetilde{\mathcal{U}}$ as the same data structure as $\widetilde{\mathcal{X}}$, with all their elements being 0.

**Do**:

(1) Save $(\widetilde{\mathcal{Z}}_{\mathrm{prev}}, \widetilde{\mathcal{U}}_{\mathrm{prev}}) \leftarrow (\widetilde{\mathcal{Z}}, \widetilde{\mathcal{U}})$.

(2) Update $\widetilde{\mathcal{X}} = \mathrm{prox}_{\eta f}(\widetilde{\mathcal{Z}}_{\mathrm{prev}} - \widetilde{\mathcal{U}}_{\mathrm{prev}})$.

(3) Update $\widetilde{\mathcal{Z}} = \mathrm{prox}_{\eta g}(\widetilde{\mathcal{X}} - \widetilde{\mathcal{U}}_{\mathrm{prev}})$.

(4) Update $\widetilde{\mathcal{U}} = \widetilde{\mathcal{U}}_{\mathrm{prev}} + \widetilde{\mathcal{X}} - \widetilde{\mathcal{Z}}$.

  **Until**: $\|\widetilde{\mathcal{U}} - \widetilde{\mathcal{U}}_{\mathrm{prev}}\| < \epsilon$, $\|\widetilde{\mathcal{Z}} - \mathcal{Z}_{\mathrm{prev}}\| < \epsilon$.

---

In Algorithm 1, the parameter $\eta$ defines the step size. The proximal operator is defined as $\text{prox}_h(\mathcal{X}) = \text{argmin}_\mathcal{Y}\left(h(\mathcal{Y}) + \frac{1}{2}\|\text{vec}(\mathcal{X} - \mathcal{Y})\|_2^2\right)$. Two essential steps of Algorithm 1 are evaluating the proximal operators for $\eta f$ and $\eta g$ in Steps (2) and (3). As $f$ is the summation of multiple terms involving non-overlapping tensors $\mathcal{X}_i^{(j)}$'s, its proximal operator can be expressed using the proximal operators of individual $p_{i,j}(\cdot)$'s, as indicated by the separable property of the proximal operator [13],

$$\text{prox}_h(\mathbf{x}_1, \mathbf{x}_2) = \left(\text{prox}_{h_1}(\mathbf{x}_1), \text{prox}_{h_2}(\mathbf{x}_2)\right)$$
$$\text{if } h(\mathbf{x}_1, \mathbf{x}_2) = h_1(\mathbf{x}_1) + h_2(\mathbf{x}_2).$$

Using the separable property again, the proximal operator of each $p_{i,j}\left(\mathcal{X}_i^{(j)}\right)$ can be expressed via the proximal operators of quadratic functions, norms, and other simple functions, whose closed-forms are available [13]. As for $g(\widetilde{\mathcal{X}})$, the separable property of the proximal operator $I_C$ can be invoked again, by noting that

$$I_C(\widetilde{\mathcal{X}}) = \sum_{i_1,\ldots,i_d} I_{C_{i_1,\ldots,i_d}}\left(\widetilde{\mathcal{X}}_1(i_1,\ldots,i_d),\ldots,\widetilde{\mathcal{X}}_m(i_1,\ldots,i_d)\right),$$

where

$C_{i_1,\ldots,i_d} = \left\{\mathcal{X}_i^{(1)}(i_1,\ldots,i_d) = \cdots = \mathcal{X}_i^{(n_i)}(i_1,\ldots,i_d), \forall\ i \in [m]\right\} \cap \left\{\mathcal{M}(i_1,\ldots,i_d) = \sum_{i=1}^m \mathcal{X}_i^{(1)}(i_1,\ldots,i_d)\right\}$

and $\widetilde{\mathcal{X}}_i(i_1,\ldots,i_d) = \left(\mathcal{X}_i^{(1)}(i_1,\ldots,i_d),\ldots,\mathcal{X}_i^{(n_i)}(i_1,\ldots,i_d)\right)$.

Each set $C_{i_1,\ldots,i_d}$ is an affine subset within $\mathbb{R}^{\sum_{i=1}^m n_i}$ defined by a system of linear equations:

$C_{i_1,\ldots,i_d} =$
$\left\{\left(\mathcal{X}_1^{(1)}(i_1,\ldots,i_d),\ldots,\mathcal{X}_1^{(n_1)}(i_1,\ldots,i_d),\ldots,\mathcal{X}_m^{(n_m)}(i_1,\ldots,i_d)\right) \in \right.$
$\mathbb{R}^{\sum_{i=1}^m n_i} \mid \mathcal{X}_1^{(1)}(i_1,\ldots,i_d) = \cdots =$
$\mathcal{X}_i^{(n_i)}(i_1,\ldots,i_d), \mathcal{M}(i_1,\ldots,i_d) =$
$\left. \sum_{i=1}^m \mathcal{X}_i^{(1)}(i_1,\ldots,i_d),\ \ i \in [m]\right\}$

denoting the space of the element $(i_1,\ldots,i_d)$ of all tensors $\mathcal{X}_i^{(j)}, i \in [m], j \in [n_i]$. The proximal operator of $I_{C_{i_1,\ldots,i_d}}$ is thus a projection onto $C_{i_1,\ldots,i_d}$ that can be evaluated using Proposition 1 (also given in reference [13]).

**Proposition 1** Let $\Omega = \{\mathbf{x} \mid \mathbf{Ax} = \mathbf{b}\}$. The proximal operator of $\eta I_\Omega(\mathbf{x})$ is given by: $\text{prox}_{\eta I_\Omega}(\mathbf{x}) = \text{proj}_\Omega(\mathbf{x}) = \mathbf{x} - \mathbf{A}^\top(\mathbf{A}^\top \mathbf{A})^{-1}(\mathbf{Ax} - \mathbf{b})$, where $\text{proj}_S(\mathbf{x}) = \text{argmin}_{\mathbf{y} \in S}\|\mathbf{x} - \mathbf{y}\|_2$.

Note that the input size of each function $I_{C_{i_1,\ldots,i_d}}$ is $\sum_{i=1}^m n_i$, which is the total number of structural properties of all components. This is generally a small number and is irrelevant to the total number of elements in the tensor. Therefore, the size of $\mathbf{A}$ is not too big. Also, the proximal operators of $I_{C_{i_1,\ldots,i_d}}$ involves the same matrix $\mathbf{A}$ for any element $(i_1,\ldots,i_d)$. Therefore, the matrix $\mathbf{A}^\top(\mathbf{A}^\top \mathbf{A})^{-1}$ only needs to be calculated once. Finally, the proximal operators of all $I_{C_{i_1,\ldots,i_d}}$'s can be evaluated in parallel. These three features significantly boost computational speed.

Now, let us revisit the two examples in the previous section and give the solution procedure for these specific problems.

***The solution to Example 1.*** To solve Problem (2), we define $\widetilde{\mathcal{X}} = \left(\mathcal{X}_1^{(1)}, \mathcal{X}_1^{(2)}, \mathcal{X}_2^{(1)}, \mathcal{X}_2^{(2)}\right)$. The functions of $f$ and $g$ in Problem (2) are

$$f(\widetilde{\mathcal{X}}) = \lambda_{1,1} p_{1,1}(\mathcal{X}_1^{(1)}) + \lambda_{1,2} p_{1,2}(\mathcal{X}_1^{(2)}) + \lambda_{2,1} p_{2,1}(\mathcal{X}_2^{(1)}) + \lambda_{2,2} p_{2,2}(\mathcal{X}_2^{(2)});$$
$$g(\widetilde{\mathcal{X}}) = I_{\mathcal{X}_1^{(1)} = \mathcal{X}_1^{(2)},\ \mathcal{X}_2^{(1)} = \mathcal{X}_2^{(2)},\ \mathcal{X}_1^{(1)} + \mathcal{X}_2^{(1)} = \mathcal{M}}(\widetilde{\mathcal{X}}).$$

The proximal operators of $p_{1,1}$, $p_{1,2}$, $p_{2,1}$, $p_{2,2}$ and $g$ are evaluated using the following procedure.

If $p(\mathbf{x}) = \|\mathbf{Ax}\|_2^2$, then the proximal operator of $\eta p(\cdot)$ is $\text{prox}_{\eta p(\cdot)}(\mathbf{x}) = (2\eta \mathbf{A}^\top \mathbf{A} + \mathbf{I})^{-1}\mathbf{x}$. Therefore, the proximal operators associated to the smoothness penalization $p_{1,1}, p_{1,2}$, and $p_{2,1}$ can be respectively expressed as

$$[\text{prox}_{\eta \lambda_{i,j} p_{i,j}}(\mathcal{X})]_{(s)} = \left(2\lambda_{i,j}\eta \mathbf{D}_s^{\top} \mathbf{D}_s^t + \mathbf{I}\right)^{-1}\mathcal{X}_{(s)}. \quad (5)$$

The proximal operator of $p_{2,2}$ can be evaluated using

$$\text{prox}_{\eta \lambda_{2,2} p_{2,2}}(\mathcal{X}) = (\mathcal{X} - \lambda_{2,2}\eta \mathcal{J})_+, \quad (6)$$

where $\mathcal{J} \in \mathbb{R}^{I_1 \times I_2 \times I_3}$ is a tensor with all elements being 1's.

The calculation of $\left(\mathcal{Y}_1^{(1)}, \mathcal{Y}_1^{(2)}, \mathcal{Y}_2^{(1)}, \mathcal{Y}_2^{(2)}\right) = \text{prox}_{\eta g}\left[\mathcal{X}_1^{(1)}, \mathcal{X}_1^{(2)}, \mathcal{X}_2^{(1)}, \mathcal{X}_2^{(2)}\right]$ is performed for each set of corresponding elements of $\mathcal{Y}_1^{(1)}, \mathcal{Y}_1^{(2)}, \mathcal{Y}_2^{(1)}, \mathcal{Y}_2^{(2)}$. Specifically, the $(i_1, i_2, i_3)$ element is updated via

$$\left[\mathcal{Y}_1^{(1)}(i_1, i_2, i_3), \mathcal{Y}_1^{(2)}(i_1, i_2, i_3), \mathcal{Y}_2^{(1)}(i_1, i_2, i_3), \mathcal{Y}_2^{(2)}(i_1, i_2, i_3)\right]^\top = \mathbf{x} - \mathbf{A}^\top(\mathbf{A}^\top \mathbf{A})^{-1}(\mathbf{Ax} - \mathbf{b}), \quad (7)$$

where

$$\mathbf{x} = \left(\mathcal{X}_1^{(1)}(i_1, i_2, i_3),\ldots,\mathcal{X}_2^{(2)}(i_1, i_2, i_3)\right)^\top, \quad \mathbf{b} = \begin{bmatrix} 0 \\ 0 \\ \mathcal{M}(i_1, i_2, i_3) \end{bmatrix},$$
$$\mathbf{A} = \begin{bmatrix} 1 & -1 & 0 & 0 \\ 0 & 0 & 1 & -1 \\ 1 & 0 & 1 & 0 \end{bmatrix}.$$

Here matrix $\mathbf{I}$ represents the identity matrices of appropriate dimension.

The optimization problem (2) can therefore be solved using the ADMM algorithm listed in Algorithm 2.

---

**Algorithm 2** ADMM algorithm for Example 1

Initialize $\widetilde{\mathcal{Z}} = \left(\mathcal{Z}_1^{(1)}, \mathcal{Z}_1^{(2)}, \mathcal{Z}_2^{(1)}, \mathcal{Z}_2^{(2)}\right)$ and $\widetilde{\mathcal{U}} = \left(\mathcal{U}_1^{(1)}, \mathcal{U}_1^{(2)}, \mathcal{U}_2^{(1)}, \mathcal{U}_2^{(2)}\right)$ with the same data structure as $\widetilde{\mathcal{X}} = \left(\mathcal{X}_1^{(1)}, \mathcal{X}_1^{(2)}, \mathcal{X}_2^{(1)}, \mathcal{X}_2^{(2)}\right)$. Set all their elements to 0.

**Do:**
(1) Save $(\widetilde{\mathcal{Z}}_{\text{prev}}, \widetilde{\mathcal{U}}_{\text{prev}}) \leftarrow (\widetilde{\mathcal{Z}}, \widetilde{\mathcal{U}})$.
(2) Update all mode-2 fibers in $\mathcal{X}_1^{(1)}$, all mode-3 fibers in $\mathcal{X}_1^{(1)}$ and all mode-1 fibers in $\mathcal{X}_2^{(1)}$ in For loops (2a-2c) and assign each element of $\mathcal{X}_2^{(2)}$ in (2d) in parallel.
    (2a) **For** all $i \in [I_1]$ $k \in [I_3]$:
        $\mathcal{X}_1^{(1)}(i, :, k) = \left(2\lambda_{1,1}\eta \mathbf{D}_2^\top \mathbf{D}_2^t + \mathbf{I}_1^{(1)}\right)^{-1}\left(\mathcal{Z}_{1,\text{prev}}^{(1)}(i, :, k) - \mathcal{U}_{1,\text{prev}}^{(1)}(i, :, k)\right)$, where

$\mathbf{D}_2^1 \in \mathbb{R}^{(I_2-1)\times I_2}$ is the first-order difference matrix and $\mathbf{I}_1^{(1)} \in \mathbb{R}^{I_2 \times I_2}$ is the identity matrix;

(2b) **For all** $i \in [I_1], k \in [I_2]$:

$\mathcal{X}_1^{(2)}(i,j,:) = \left(2\lambda_{1,2}\eta \mathbf{D}_3^{1\top}\mathbf{D}_3^1 + \mathbf{I}_1^{(2)}\right)^{-1}\left(\mathcal{Z}_{1,\text{prev}}^{(2)}(i,j,:) - \mathcal{U}_{1,\text{prev}}^{(2)}(i,j,:)\right)$ , where $\mathbf{D}_3^1 \in \mathbb{R}^{(I_3-1)\times I_3}$ is the first-order difference matrix and $\mathbf{I}_1^{(2)} \in \mathbb{R}^{I_3 \times I_3}$ is the identity matrix;

(2c) **For all** $j \in [I_2], k \in [I_3]$:

$\mathcal{X}_2^{(1)}(:,j,k) = \left(2\lambda_{1,2}\eta \mathbf{D}_1^{2\top}\mathbf{D}_1^2 + \mathbf{I}_2^{(1)}\right)^{-1}\left(\mathcal{Z}_{2,\text{prev}}^{(1)}(:,j,k) - \mathcal{U}_{2,\text{prev}}^{(1)}(:,j,k)\right)$ , where $\mathbf{D}_1^2 \in \mathbb{R}^{(I_1-1)\times I_1}$ is the second-order difference matrix and $\mathbf{I}_2^{(1)} \in \mathbb{R}^{I_1 \times I_1}$ is the identity matrix;

(2d) $\mathcal{X}_2^{(2)} = \left(\mathcal{Z}_{2,\text{prev}}^{(2)} - \mathcal{U}_{2,\text{prev}}^{(2)} - \lambda_{2,2}\eta \mathcal{J}\right)_+$.

(2e) $\tilde{\mathcal{X}} \leftarrow \left(\mathcal{X}_1^{(1)}, \mathcal{X}_1^{(2)}, \mathcal{X}_2^{(1)}, \mathcal{X}_2^{(2)}\right)$.

(3) **For all** $(i_1, i_2, i_3)$:

$\left(\mathcal{X}_1^{(1)}(i_1,i_2,i_3), \dots, \mathcal{Z}_2^{(2)}(i_1,i_2,i_3)\right) \leftarrow \mathbf{x} - \mathbf{A}^\top(\mathbf{A}^\top\mathbf{A})^{-1}(\mathbf{A}\mathbf{x} - \mathbf{b})$,

where

$\mathbf{x} = \left(\mathcal{X}_1^{(1)}(i_1,i_2,i_3) - \mathcal{U}_1^{(1)}(i_1,i_2,i_3), \dots, \mathcal{X}_2^{(2)}(i_1,i_2,i_3) - \mathcal{U}_2^{(2)}(i_1,i_2,i_3)\right)$

and

$\mathbf{A} = \begin{bmatrix} 1 & -1 & 0 & 0 \\ 0 & 0 & 1 & -1 \\ 1 & 0 & 1 & 0 \end{bmatrix}, \mathbf{b} = \begin{bmatrix} 0 \\ 0 \\ \mathcal{M}(i_1,i_2,i_3) \end{bmatrix}$.

(4) Update $\tilde{\mathcal{U}} = \tilde{\mathcal{U}}_{\text{prev}} + \tilde{\mathcal{X}} - \tilde{\mathcal{Z}}$.

**Until:** $\|\tilde{\mathcal{U}} - \tilde{\mathcal{U}}_{\text{prev}}\| < \epsilon$, $\|\tilde{\mathcal{Z}} - \mathcal{Z}_{\text{prev}}\| < \epsilon$.

**The solution to Example 2.** First, we define the copies $\tilde{\mathcal{X}} = \left(\mathcal{X}_1^{(1)}, \mathcal{X}_1^{(2)}, \mathcal{X}_1^{(3)}, \mathcal{X}_2^{(1)}, \mathcal{X}_2^{(2)}, \mathcal{X}_3^{(1)}\right)$. The problem is then transformed into the canonical formulation (4), where $f(\tilde{\mathcal{X}}) = \lambda_{1,1}p_{1,1}(\mathcal{X}_1^{(1)}) + \lambda_{1,2}p_{1,2}(\mathcal{X}_1^{(2)}) + \lambda_{1,3}p_{1,3}(\mathcal{X}_1^{(3)}) + \lambda_{2,1}p_{2,1}(\mathcal{X}_2^{(1)}) + \lambda_{2,2}p_{2,2}(\mathcal{X}_2^{(2)}) + \lambda_{3,1}p_{3,1}(\mathcal{X}_3^{(1)})$ and $g(\tilde{\mathcal{X}}) = I_{\mathcal{X}_1^{(1)}=\mathcal{X}_1^{(2)}=\mathcal{X}_1^{(3)}, \mathcal{X}_2^{(1)}=\mathcal{X}_2^{(2)}, \mathcal{X}_1^{(1)}+\mathcal{X}_2^{(1)}+\mathcal{X}_3^{(1)}=\mathcal{M}}(\tilde{\mathcal{X}})$.

To perform the ADMM algorithm, we need to evaluate the proximal operators of all $p_{i,j}$'s and $g$.

The proximal operators of $p_{1,1}$ and $p_{1,2}$ are evaluated in the same way as that $p_{1,1}$ and $p_{1,2}$ in Example 1.

$$[\text{prox}_{\eta\lambda_{i,j}p_{i,j}}(\mathcal{X})]_{(s)} = \left(2\lambda_{i,j}\eta \mathbf{D}_s^\top\mathbf{D}_s^1 + \mathbf{I}\right)^{-1}\mathcal{X}_{(s)}. \quad (8)$$

The proximal operators of $p_{1,3}$ and $p_{2,1}$ are in the form of the nuclear norm, whose proximal operator is given by

$$\text{prox}_{\eta\|\cdot\|_*}(\mathbf{A}) = \sum_i (\sigma_i - \eta)_+ \mathbf{u}_i\mathbf{v}_i^\top, \quad (9)$$

for which $\mathbf{A} = \sum_i \sigma_i \mathbf{u}_i\mathbf{v}_i^\top$ is the singular value decomposition of $\mathbf{A}$ [13].

The proximal operators of functions $p_{2,2}$ and $p_{3,1}$ are in the form of $\ell_1$-norm. Their proximal operators are given by

$$\text{prox}_{\eta\|\cdot\|_1}(\mathcal{X}) = (\mathcal{X} - \eta)_+. \quad (10)$$

The operation $\text{prox}_{\eta g}$ is again calculated on the groups of the same elements in the five input tensors by Proposition 1. The calculation of $\left(\mathcal{Y}_1^{(1)}, \mathcal{Y}_1^{(1)}, \mathcal{Y}_1^{(3)}, \mathcal{Y}_2^{(1)}, \mathcal{Y}_2^{(2)}, \mathcal{Y}_3^{(1)}\right) = \text{prox}_{\eta g}\left[\mathcal{X}_1^{(1)}, \mathcal{X}_1^{(2)}, \mathcal{X}_1^{(3)}, \mathcal{X}_2^{(1)}, \mathcal{X}_2^{(2)}, \mathcal{X}_3^{(1)}\right]$ is performed for each set of corresponding elements of $\mathcal{Y}_1^{(1)}, \mathcal{Y}_1^{(3)}, \mathcal{Y}_1^{(3)}, \mathcal{Y}_2^{(1)}, \mathcal{Y}_2^{(2)}, \mathcal{Y}_3^{(1)}$. The $(i_1, i_2, i_3)$ element is updated via

$$\begin{aligned} &\left[\mathcal{Y}_1^{(1)}(i_1,i_2,i_3), \mathcal{Y}_1^{(2)}(i_1,i_2,i_3), \mathcal{Y}_1^{(3)}(i_1,i_2,i_3), \right.\\ &\left.\mathcal{Y}_2^{(1)}(i_1,i_2,i_3), \mathcal{Y}_2^{(2)}(i_1,i_2,i_3), \mathcal{Y}_3^{(1)}(i_1,i_2,i_3)\right]^\top \\ &\quad = \mathbf{x} - \mathbf{A}^\top(\mathbf{A}^\top\mathbf{A})^{-1}(\mathbf{A}\mathbf{x} - \mathbf{b}), \end{aligned} \quad (11)$$

with $\mathbf{x} = \left(\mathcal{X}_1^{(1)}(i_1,i_2,i_3), \dots, \mathcal{X}_3^{(1)}(i_1,i_2,i_3)\right)^\top$,

$$\mathbf{A} = \begin{bmatrix} 1 & -1 & 0 & 0 & 0 & 0 \\ 0 & 1 & -1 & 0 & 0 & 0 \\ 0 & 0 & 0 & 1 & -1 & 0 \\ 1 & 0 & 0 & 1 & 0 & 1 \end{bmatrix} \text{ and } \mathbf{b} = \begin{bmatrix} 0 \\ 0 \\ 0 \\ \mathcal{M}(i_1,i_2,i_3) \end{bmatrix}.$$

The canonical form (3) can be solved using the ADMM algorithm listed in Algorithm 3.

---

**Algorithm 3** ADMM algorithm for Example 2

Initialize $\tilde{\mathcal{Z}} = \left(\mathcal{Z}_1^{(1)}, \mathcal{Z}_1^{(2)}, \mathcal{Z}_1^{(3)}, \mathcal{Z}_2^{(1)}, \mathcal{Z}_2^{(2)}, \mathcal{Z}_3^{(1)}\right)$ and $\tilde{\mathcal{U}} = \left(\mathcal{U}_1^{(1)}, \mathcal{U}_1^{(2)}, \mathcal{U}_1^{(3)}, \mathcal{U}_2^{(1)}, \mathcal{U}_2^{(2)}, \mathcal{U}_3^{(1)}\right)$ as the same data structure as $\tilde{\mathcal{X}} = \left(\mathcal{X}_1^{(1)}, \mathcal{X}_1^{(2)}, \mathcal{X}_1^{(3)}, \mathcal{X}_2^{(1)}, \mathcal{X}_2^{(2)}, \mathcal{X}_3^{(1)}\right)$, with all their elements being 0.

**Do**:

(1) Save $(\tilde{\mathcal{Z}}_{\text{prev}}, \tilde{\mathcal{U}}_{\text{prev}}) \leftarrow (\tilde{\mathcal{Z}}, \tilde{\mathcal{U}})$.

(2) Update all mode-2 fibers in $\mathcal{X}_1^{(1)}$, all mode-3 fibers in $\mathcal{X}_1^{(2)}$, all elements of $\mathcal{X}_2^{(2)}$, all elements of $\mathcal{X}_3^{(1)}$ in parallel with $\mathcal{X}_{1(1)}^{(3)}$ and $\mathcal{X}_{2(1)}^{(1)}$ in (2a)-(2e) below.

(2a) For all $i \in [I_1]$ $k \in [I_3]$:

$\mathcal{X}_1^{(1)}(i,:,k) = \left(2\lambda_{1,1}\eta \mathbf{D}_2^{1\top}\mathbf{D}_2^1 + \mathbf{I}_1^{(1)}\right)^{-1}\left(\mathcal{Z}_{1,\text{prev}}^{(1)}(i,:,k) - \mathcal{U}_{1,\text{prev}}^{(1)}(i,:,k)\right)_{(2)}$, where $\mathbf{D}_2^1 \in \mathbb{R}^{(I_2-1)\times I_2}$ is the first-order difference matrix and $\mathbf{I}_1^{(1)} \in \mathbb{R}^{I_2 \times I_2}$ is the identity matrix;

(2b) **For all** $i \in [I_1], j \in [I_2]$:

$\mathcal{X}_1^{(2)}(i,j,:) = \left(2\lambda_{1,2}\eta \mathbf{D}_3^{1\top}\mathbf{D}_3^1 + \mathbf{I}_1^{(1)}\right)^{-1}\left(\mathcal{Z}_{1,\text{prev}}^{(2)}(i,j,:) - \mathcal{U}_{1,\text{prev}}^{(2)}(i,j,:)\right)_{(3)}$, where $\mathbf{D}_3^1 \in \mathbb{R}^{(I_3-1)\times I_3}$ is the first-order difference matrix and $\mathbf{I}_1^{(2)} \in \mathbb{R}^{I_3 \times I_3}$ is the identity matrix;

(2c) $\mathcal{X}_{1(1)}^{(3)} = \sum_i (\sigma_i - \eta)_+ \mathbf{u}_i\mathbf{v}_i^\top$, where $\sum_i \sigma_i \mathbf{u}_i\mathbf{v}_i^\top$ is the singular value decomposition of $\left(\mathcal{Z}_{1,\text{prev}}^{(3)} - \mathcal{U}_{1,\text{prev}}^{(3)}\right)_{(1)}$;

(2d) $\mathcal{X}_{2(1)}^{(1)} = \sum_i (\sigma_i - \eta)_+ \mathbf{u}_i\mathbf{v}_i^\top$, where $\sum_i \sigma_i \mathbf{u}_i\mathbf{v}_i^\top$ is the singular value decomposition of $\left(\mathcal{Z}_{2,\text{prev}}^{(1)} - \mathcal{U}_{2,\text{prev}}^{(1)}\right)_{(1)}$;

(2e) $\mathcal{X}_2^{(2)} = \left(\mathcal{Z}_{2,\text{prev}}^{(2)} - \mathcal{U}_{2,\text{prev}}^{(2)} - \eta\right)_+$ and $\mathcal{X}_3^{(1)} = \left(\mathcal{Z}_{3,\text{prev}}^{(1)} - \mathcal{U}_{3,\text{prev}}^{(1)} - \eta\right)_+$.

(3) For all $(i_1, i_2, i_3)$:

$$\left(\mathcal{Z}_1^{(1)}(i_1,i_2,i_3),\dots,\mathcal{Z}_3^{(1)}(i_1,i_2,i_3)\right)$$
$$\leftarrow \mathbf{x} - \mathbf{A}^\top(\mathbf{A}^\top\mathbf{A})^{-1}(\mathbf{A}\mathbf{x}-\mathbf{b}),$$

where

$$\mathbf{x} = \left(\mathcal{X}_1^{(1)}(i_1,i_2,i_3) - \mathcal{U}_1^{(1)}(i_1,i_2,i_3),\dots,\mathcal{X}_3^{(1)}(i_1,i_2,i_3) - \mathcal{U}_3^{(1)}(i_1,i_2,i_3)\right),$$

$$\mathbf{A} = \begin{bmatrix} 1 & -1 & 0 & 0 & 0 & 0 \\ 0 & 1 & -1 & 0 & 0 & 0 \\ 0 & 0 & 0 & 1 & -1 & 0 \\ 1 & 0 & 0 & 1 & 0 & 1 \end{bmatrix}, \quad \text{and} \quad \mathbf{b} = \begin{bmatrix} 0 \\ 0 \\ 0 \\ \mathcal{M}(i_1,i_2,i_3) \end{bmatrix}.$$

(4) Update $\widetilde{\mathcal{U}} = \widetilde{\mathcal{U}}_{\text{prev}} + \widetilde{\mathcal{X}} - \widetilde{\mathcal{Z}}$.

**Until:** $\|\widetilde{\mathcal{U}} - \widetilde{\mathcal{U}}_{\text{prev}}\| < \epsilon$, $\|\widetilde{\mathcal{Z}} - \mathcal{Z}_{\text{prev}}\| < \epsilon$.

### C. Discussion

It can be seen from those two examples that the ATD framework successfully integrates multiple structural information required for every tensor component. It is both a unification and an extension of the existing decomposition methods, such as the SSD, the STSSD, and the RPCA. In particular, when we ignore the temporal smoothness property to the crack in Example 1, the formulation will be reduced to the SSD method. If we invoke the temporal smoothness property to the background, we will arrive at the ST-SSD method. However, neither the SSD nor the ST-SSD fully describes the setup of Example 1, because neither of them incorporates the information that the crack is smooth in time but sparse in space and that the background is smooth in space but may not be smooth in time. In Example 2, if we did not distinguish the static anomaly and static background, the problem formulation would be reduced to the RPCA for foreground detection. However, the RPCA cannot fully describe the setup of Example 2 because it does not incorporate the sparsity information of static hotspots or smoothness information of the background.

We can also see from solution algorithms of those two examples that the adopted ADMM algorithm is highly parallelable. For example, in steps 2(a-c) of Algorithm 2, the update of each tensor component can be done fiber by fiber; In step 3, the update of each element can be performed in parallel. This operation can be distributed on several processors which is highly suitable for tensor with high dimensions.

In the following subsections, we first discuss the selection of tuning parameters and then analyze the non-unique solution issue: what to do if the decomposition cannot be performed.

### D. Selection of tuning parameters

In the general formulation (1), the tuning parameters need to be specified. A large value of the tuning parameter $\lambda_{i,j}$ tends to enhance the corresponding penalty terms $p_{i,j}(\cdot)$.

If some training samples containing one or more existing tensor data $\mathcal{X}_{\{1\}},\dots,\mathcal{X}_{\{n\}}$ and their real additive components of interest $\mathcal{X}_{\{1\}i},\dots,\mathcal{X}_{\{n\}i}, i \in [m]$ are available, the optimal tuning parameters can be determined by minimizing the error [16]:

$$\hat{\boldsymbol{\lambda}} = \operatorname{argmin}_{\boldsymbol{\lambda}} \left\{ \sum_{i=1}^m \sum_{j=1}^n L_i \left( \mathcal{X}_{\{j\}i} - \widehat{\mathcal{X}_{\{j\}i}}(\boldsymbol{\lambda}) \right) \right\}.$$

Here $\widehat{\mathcal{X}_{\{j\}i}}(\boldsymbol{\lambda})$ is the estimation of the $i$th tensor component by solving the problem involving the $j$th training tensor $\mathcal{X}_{\{j\}}$ and using the tuning parameters $\boldsymbol{\lambda}$. $L_i(\cdot)$ specifies the loss involving the difference between the true component $\mathcal{X}_{\{1\}}$ and the estimated component $\widehat{\mathcal{X}_{\{1\}}}(\boldsymbol{\lambda})$. The loss function $L_i(\cdot)$'s can be selected depending on specific scenarios: if we only care about the nonzero part in some components, as in the case of anomaly detections, we should use 0-1 loss function $L_i(\mathcal{X}) = \|\text{vec}(\mathcal{X})\|_0$; If we care about the value of the extract features, we may prefer to use the Frobenious loss: $L_i(\mathcal{X}) = \|\mathcal{X}\|_F$. If the decomposition result of the $i$th component is not of interest, the loss function of $L_i(\cdot)$ can be set to 0.

If we do not have a training sample, $\boldsymbol{\lambda}$ should be selected empirically based on the structural assumption we want to incorporate. It is also suggested that one tuning parameter is adjusted at a time based on the result of decomposition, until finding the values that lead to the optimal result. In Example 1, the decomposed background image generated by larger values of $\lambda_{1,1}$ and $\lambda_{1,2}$ tends to be smoother than that generated by small values of $\lambda_{1,1}$ and $\lambda_{1,2}$. In other words, large values of $\lambda_{1,1}$ and $\lambda_{1,2}$ lead to smooth background image. The changes among extracted crack images generated by a large value of $\lambda_{2,1}$ tends to be smoother, and the extracted crack images generated by a large value of $\lambda_{3,1}$ tends to be sparser. Therefore, we select the tuning parameters in this application using the following procedure: we first select $\lambda_{1,1}$, $\lambda_{1,2}$, and $\lambda_{3,1}$ because the selected tensor components are very sensitive to the value $\frac{\lambda_{3,1}}{\lambda_{1,1}+\lambda_{1,2}}$. Then, the tuning parameters for temporal smoothness penalty $\lambda_{2,1}$ is selected accordingly to preserve the low-intensity pixels on the crack and to shrink intensity of other pixels that belongs to the background, because the sparsity penalty tends to shrink the value of low intensity pixels in the tensor component $\mathcal{X}_2$ to zero, regardless of the low-intensity pixels are on the crack or not. In Example 2, a large $\lambda_{2,1}$ and $\lambda_{1,3}$ help to enhance the low rank property of the mode-(1) matricization of static hotspot tensor and background tensor, the extracted moving hotspots generated by a large value of $\lambda_{3,1}$ tends to be sparser, and the background generated by large values of $\lambda_{1,1}$ and $\lambda_{1,2}$ tends to be smoother; the extracted static hotspots generated by a large value of $\lambda_{2,2}$ tends to be sparser. We select the tuning parameters in Example 2 as follows. First, we select $\lambda_{2,1}, \lambda_{1,3}$, and $\lambda_{3,1}$ to separate the moving hotspots and static components, including the background and static hotspots. Then, we select $\lambda_{1,1}, \lambda_{1,2}$, and $\lambda_{2,2}$ that control the smoothness of the image in the background tensor and the sparsity of the static hotspot tensor, to separate the static hotspot from the smooth background.

### E. Problem of nonunique solution

It should be pointed out that the solution to the optimization problem sometimes may not be unique. This is because structural assumptions in the framework are not rich enough to determine each component. For example, two tensor components that have the same sparse structural property are not distinguishable under the framework. To generate a unique solution, we propose the following two solutions. One solution is to incorporate some prior knowledge that distinguishes the tensor components that cannot be uniquely identified. For instance, if we know that the nonzero elements in two sparse tensor components form different shapes, such as square and circle, we can express these components using two different sets of functional bases to ensure a unique solution. If there is no prior knowledge available, we propose to add another regularization term $\epsilon \sum_{i=1}^{d} \|\mathcal{X}_i\|_F^2$ to the objective function, where $\epsilon$ is a small number. With this term, the objective function becomes strongly convex and the unique solution exists.

## IV. SIMULATION STUDIES FOR PERFORMANCE EVALUATION

In this section, we will present our simulation studies on those two examples to illustrate the ATD framework and demonstrate its effectiveness.

### A. Simulation study for Example 1

The data for monitoring the crack growing process include $I_1 = 30$ consecutive measurement images of size $40 \times 40$. These images form a tensor $\mathcal{M} \in \mathbb{R}^{30 \times 40 \times 40}$. We simulate $\mathcal{M}$ by summing up two tensors $\mathcal{X}_1$ and $\mathcal{X}_2$ that represent the background and the crack, respectively. Each $\mathcal{X}_1(i,:,:), i \in [I_1]$, is generated by a 2D smooth Gaussian process representing the background. Most values in the image $\mathcal{X}_2(i,:,:)$ are zeros, and the non-zero values of $\mathcal{X}_2(i,:,:)$ gradually grows when the index $i$ increases from 1 to 30, representing the crack growing on the wall. These values are generated from i.i.d. $N(0.1, 0.01)$ random variables to represent the random lighting and shadowing conditions. The first images of Fig. 6(a) and Fig. 6(b) illustrates the $20_{th}$ and the $30_{th}$ images in $\mathcal{M}$, the second images in Figs. 6(a) and 6(b) illustrate the corresponding images for actual crack which is a continuous line and the third images in Figs. 6(a) and 6(b) illustrate the images for the simulated crack under irregularly illuminated conditions.

We then decompose $\mathcal{M}$ into two components $\hat{\mathcal{X}}_1$ and $\hat{\mathcal{X}}_2$ by solving Problem (2). In the ADMM algorithm, the step size is $\eta = 0.01$, and the tuning parameters are $\lambda_{1,1} = \lambda_{1,2} = 1, \lambda_{2,1} = 10$, and $\lambda_{2,2} = 0.08$. The fifth images of Figs. 6(a) and 6(b) illustrate the estimated crack of the $20_{th}$ and the $30_{th}$ image using the ATD-based method. It is shown that the ATD-based method captures the growth of the whole crack accurately. The video illustrating the result of the decomposition is provided in the supplementary material of this paper.

For comparison, we also applied the SSD method to each image in $\mathcal{M}$. The crack images obtained from the SSD method are shown in the fourth images in Figs. 6(a), 6(b). It can be seen that the SSD only captures some points on the crack because the smoothness of the crack in the temporal mode is ignored. The $\ell_1$ regularization in the SSD encourages sparsity of the anomaly but pushes all pixels with low intensity to zero, including the ones on the crack. In contrast, the ATD-based method also promotes the temporal smoothness of $\mathcal{X}_2$, which preserves the pixels with low intensity on the crack, whereas penalizes other pixels to zeros.

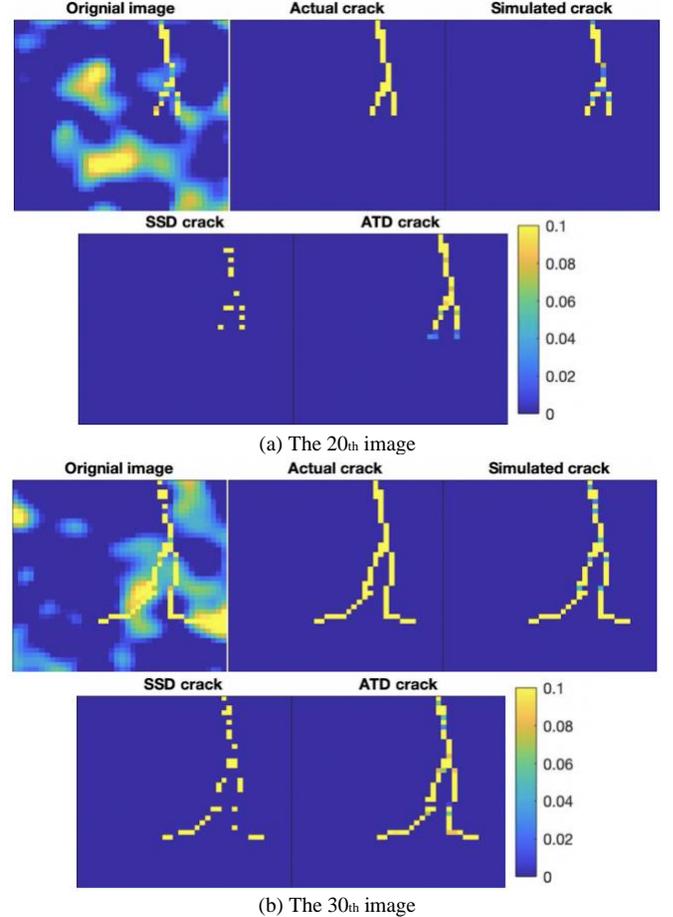

(a) The $20_{th}$ image

(b) The $30_{th}$ image

Fig. 6. Illustration of decomposed images using the ATD and the SSD methods

### B. Simulation study for Example 2

The tensor $\mathcal{M}$ in Example 2 is generated to simulate the consecutive measurements taken from a thermal camera in a heated surface monitoring process. It also contains 30 images of size $40 \times 40$, and it is generated by summing up three tensors $\mathcal{X}_1$, $\mathcal{X}_2$, and $\mathcal{X}_3$ of the same size that represents the true background, the static hotspot, and the moving hotspot respectively. Among them, each mode 1 slice of the tensor $\mathcal{X}_1$ is generated from

$$\mathcal{X}(i,:,:) = U_i \mathbf{T}_0 + (1 - U_i)\mathbf{T}_1,$$

where $\mathbf{T}_0$ is a $40 \times 40$ matrix representing the heating effect of the heating process, $\mathbf{T}_1$ is a matrix of the same size representing the cooling effect and $U_i$ is a $U[0,1]$ random variable representing a random combination of the two effects.

The images representing the matrices $\mathbf{T}_0$ and $\mathbf{T}_1$ are shown in Fig. 7. To simulate the heating effect of a single point heating source at the center of this image, we generate $\mathbf{T}_0(i,j)$ using the value of $f_0(i,j)$, where $f_0$ is the probability density function of $N((20,20)^\top, 10\mathbf{I})$, where $\mathbf{I}$ is a $2 \times 2$ identity matrix. Then, we transform all values of $\mathbf{T}_0$ linearly such that the maximum and minimum value of $\mathbf{T}_0$ are 1 and 0, respectively. With this setup, the maximum value within $\mathbf{T}_0$ is 1, located at the center of the image; when the pixel moves farther way from the center, the value of $\mathbf{T}_0(i,j)$ gradually drops to 0. To simulate the cooling effect, we generate $\mathbf{T}_1(i,j)$ using a linear function $f_1(i,j) = c_1(i+j) + c_2$, where $c_1$ and $c_2$ are adjusted so that the maximum and minimum value within the matrix $\mathbf{T}_1$ are 0 and 1 respectively. It represents that the coolant for the surface flows from the upper-left corner to the bottom-right corner of the image.

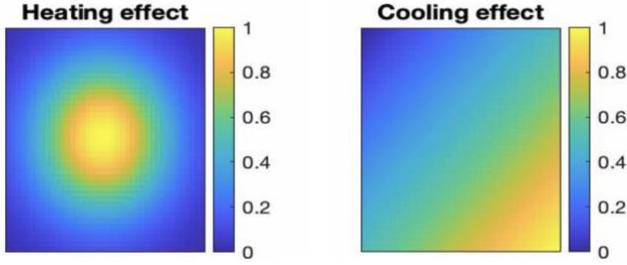

Fig. 7. Illustration of cooling and heating effect temperature field.

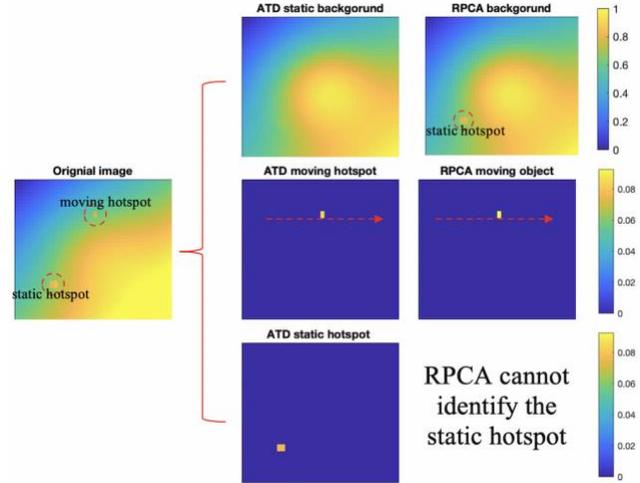

Fig. 8. Illustration of the 20th decomposed images using the ATD and the RPCA methods

Each image within the tensor $\mathcal{X}_2$ are the same, and the non-zero values in these images are located in a fixed $2 \times 2$ rectangle with intensity value 1 in their lower-left corners. The non-zero values in each image of the tensor $\mathcal{X}_3$ are also located in a $2 \times 2$ rectangle with intensity value 1. However, this rectangle locates on the upper-left part of the image. When the image index $i$ increases, the rectangle in $\mathcal{X}_3(i,:,:)$ moves from the left side to the right side across the images.

We decompose the tensor $\mathcal{M}$ into components $\widehat{\mathcal{X}}_1$, $\widehat{\mathcal{X}}_2$, and $\widehat{\mathcal{X}}_3$ by solving Problem (3). The step size $\eta = 0.01$. Tuning parameters are $\lambda_{1,1} = \lambda_{1,2} = 30$, $\lambda_{1,3} = \lambda_{2,1} = 1$, $\lambda_{2,2} = 1.9$, and $\lambda_{3,1} = 2$.

In Fig. 8, we illustrate the 20th simulated images, with the decomposed background, the static hotspot, and the moving hotspot using the ATD method. The decomposed background and the moving object from the RPCA method are illustrated in the third column. The video illustrating the result of the decomposition is provided in the supplementary material of this paper. It can be seen that the proposed ATD method successfully separates the static background, the static hotspot, and the moving hotspot. Comparing the result of the ATD, the RPCA can only identify the moving object but fails to separate the true background and the static hotspot. This is because the RPCA does not consider the spatial smoothness of the background.

From the two simulation studies, we verify that the ATD framework provides greater capability of separating the tensor of interest into multiple components than the existing methods by considering richer structural properties of the components in individual slices.

## V. A CASE STUDY WITH THE 3D MEDICAL IMAGE ANALYSIS

In this section, we revisit the case study in the Introduction: medical image analysis for three-dimensional (3D) printed phantoms [6], to further demonstrate the advantage of the ATD method.

TAVR is an alternative option for aortic stenosis patients with high surgical risk to perform aortic valve replacement [6]. AVC has been proposed as an important determinant for PVR of TAVR. To avoid PVR, the physicians use multi-material 3D printing to fabricate a prototype for patients' aortic root anatomies, upon which they plan the surgery. The parts representing the heart tissues and the AVC regions in the prototype are manufactured with materials of different mechanical properties for accurate simulation. Characterizing the shape of the AVC regions based on the patients' CT scanning images is critical for fabricating a geometrically-accurate prototype, and this task is conventionally accomplished by a board-certified cardiologist [6]. In this case study, we formulate a problem using the ATD framework, aiming at extracting the AVC regions automatically.

From the pre-procedural contrast-enhanced CT scan of some patients' aortic region, we select 12 images of size $101 \times 101$ that are collected sequentially for the aortic value. These images form a tensor $\mathcal{M}$ of size $12 \times 101 \times 101$. Each image shall be divided into 3 regions, including a contrast-enhanced blood pool with moderate intensity, the soft tissues with low intensity, and the AVC regions with high intensity. We focus on separating AVC regions from the blood pool and the soft tissues.

Specifically, the tensor $\mathcal{M}$ is decomposed into three components: the background of the blood pool and the soft tissues $\mathcal{X}_1$, the AVC regions $\mathcal{X}_2$, and the measurement error $\mathcal{X}_3$. Anatomic structure of the heart indicates that the background images are smooth, but they are different because the outer profiles of the aorta are not the same. Therefore, the regularization of smoothness is applied to every image of $\mathcal{X}_1$. The AVC regions are small regions attached to the inner side of the aorta and they do not change much between consecutive cross-sections. Therefore, the smoothness regularization in time and the sparsity regularization in space is applied to $\mathcal{X}_2$. The measurement error generally takes small values, and thus squared-$\ell_2$ norm is applied to $\mathcal{X}_3$. The problem formulation is given as follows:

$$\underset{\mathcal{X}_1, \mathcal{X}_2, \mathcal{X}_3}{\text{minimize}} \; \lambda_{1,1}\left(\left\|\mathbf{D}_1 \mathcal{X}_{1(2)}\right\|_F^2 + \left\|\mathbf{D}_1 \mathcal{X}_{1(3)}\right\|_F^2\right) +$$
$$\lambda_{2,1}\left\|\mathbf{D}_1 \mathcal{X}_{2(1)}\right\|_F^2 + \lambda_{2,2}\|\text{vec}(\mathcal{X}_2)\|_1 + \lambda_{3,1}\|\mathcal{X}_3\|_2^2,$$
$$\text{subject to } \mathcal{M} = \mathcal{X}_1 + \mathcal{X}_2 + \mathcal{X}_3.$$

With tuning parameters $\lambda_{1,1} = 10$, $\lambda_{2,1} = 0.7$, $\lambda_{2,2} = 0.16$, and $\lambda_{3,1} = 1$, we solved this problem using the ADMM algorithm. Fig. 9 (a) illustrates the $3\text{rd}$-$7\text{th}$ images in the solution of $\mathcal{X}_2$, reflecting the AVC regions in five consecutive cross-sections. As discussed before, for 3D printing of the prototype of patients' aortic root anatomies, we care about nonzero part in the extracted image and set all the nonzero values in $\mathcal{X}_2$ to 1. The video illustrating the result of the decomposition is provided in the supplementary material.

The result shows that the smooth change of AVC regions across different images is captured by the proposed ATD method, which reflects the anatomic reality. For comparison, we also conduct the SSD to extract the AVC regions on individual images. As shown in Fig. 9, for the $3\text{rd}$-$7\text{th}$ images, ATD methods achieve similar performance in extracting the AVC regions when the intensity of the anomaly region is high with respect to the background and noise. In the $7\text{th}$ image (see red circles), when the intensity of the AVC regions is low, the SSD method failed to fully extract the anomaly region. When we use a small tuning parameter for sparse penalty in the SSD method, much noise will be introduced into the extracted AVC regions. However, ATD method can still achieve satisfactory performance in this case, because the ATD takes the similarity of the AVC regions' locations between images into consideration, which helps to preserve the pixels with low intensity on the AVC regions, whereas setting other pixels to zeros. This case study further illustrates the versatility of the ATD framework and demonstrates its ability to solving real-world problems. It is worth to mention that the values of the extracted AVC regions using ATD method is smaller than that using the SSD method, due to the extra smoothness penalty added in the formulation of ATD method. However, it does not affect the boundary of the model, determined by the nonzero values in the decomposed images.

## VI. CONCLUSION

Tensor data becomes increasingly common in engineering applications. In this article, we propose an ATD framework to extract the quality-related information from tensor data based on the structural properties of the tensor components.

The ATD framework achieves the tensor decomposition through integrating multiple types of regularizations on the tensor components, corresponding to assorted structural information such as smoothness, sparsity, and low rank on different modes. In this framework, the corresponding

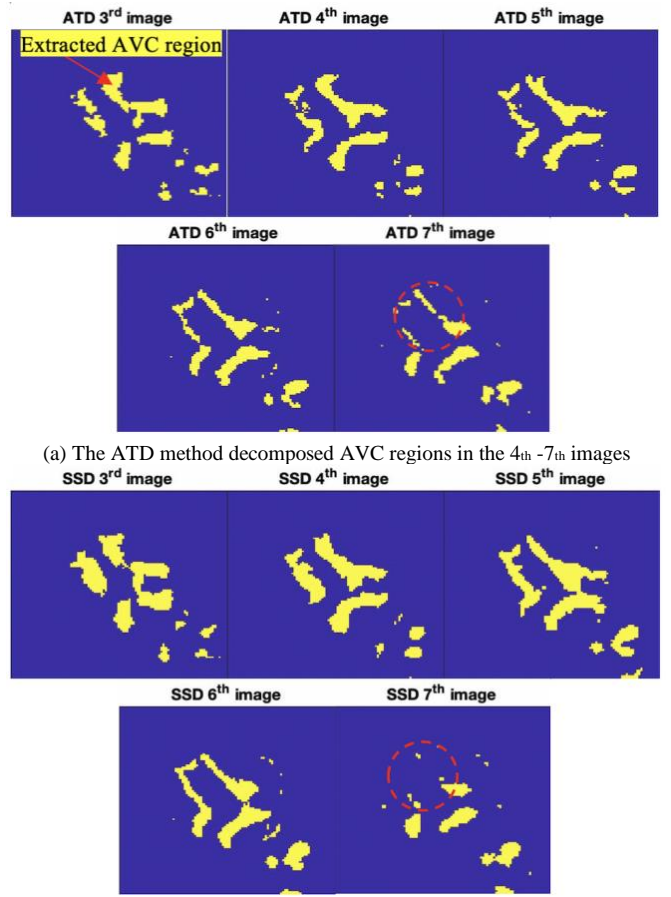

(a) The ATD method decomposed AVC regions in the $4\text{th}$ -$7\text{th}$ images

(b) The SSD method decomposed AVC regions in the $4\text{th}$ -$7\text{th}$ images
Fig. 9. Extracted AVC regions using the ATD and the SSD methods

structural property to describe the structural information is systematically defined for tensor data for the first time. As a unification and extension for the existing decomposition methods, such as SSD, STSSD, and RPCA, it provides a general framework to solve a class of tensor decomposition problem.

Computation is a major challenge of solving additive tensor decomposition problems: the number of decision variables is the number of the tensor components multiplies the number of elements in each tensor, which can be huge. To solve this large-scale problem efficiently, we adopt a highly parallelizable ADMM method.

Throughout the article, we use two examples to demonstrate the versatility of the ATD framework and illustrate its effectiveness. The case study in medical image analysis demonstrates that the ATD framework can accurately identify the AVC regions by capturing its smooth change between consecutive images. The authors believe that the ATD framework can be applied in an even wider range of applications for tensor data analysis.

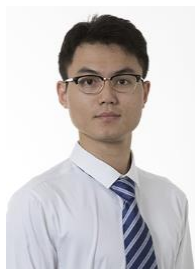

**Shancong Mou** is a Ph.D. student in the Stewart School of Industrial and Systems Engineering at Georgia Institute of Technology. He received his B.S. in Energy and Power Engineering from Xi'an Jiaotong University, Xi'an, China, in 2017. His research interests include physics-informed machine learning, medical image analysis, and data analytics for monitoring,
control and diagnostics of complex engineering systems. He is a member of IISE and INFORMS.

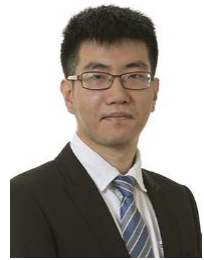

**Andi Wang** is a Ph.D. candidate in the Stewart School of Industrial and Systems Engineering at Georgia Institute of Technology. He received his B.S. in statistics from Peking University in 2012 and a Ph.D. from Hong Kong University of Science and Technology in 2016. His research interests include advanced statistical modeling, large-scale optimization, and machine learning for manufacturing and healthcare system performance improvements via process monitoring, diagnostics, prognostics, and control. He is a member of IISE and INFORMS.

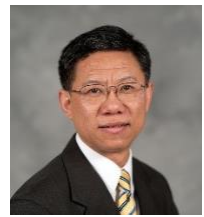

**Chuck Zhang** received the B.S. and M.S. degrees in mechanical engineering from Nanjing University of Aeronautics and Astronautics, Nanjing, China, in 1984 and 1987, respectively, an MS degree in Industrial Engineering from the State University of New York at Buffalo in 1990, and the Ph.D. degree in industrial engineering from the University of Iowa, Iowa City, IA, USA, in 1993.

Dr. Zhang is currently a Harold E. Smalley Professor with the H. Milton Stewart School of Industrial and Systems Engineering at the Georgia Institute of Technology, Atlanta, GA, USA. He has authored over 190 refereed journal articles and 210 conference papers. He also holds 24 U.S. patents. Dr. Zhang is a fellow of Institute of Industrial and Systems Engineers (IISE). His current research interests include additive manufacturing (3-D printing and printed electronics), advanced composite and nanocomposite materials manufacturing, and bio-manufacturing.

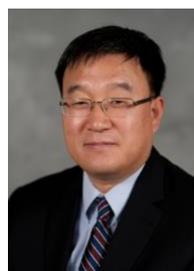

**Jianjun Shi** received the B.S. and M.S. degrees in automation from the Beijing Institute of Technology in 1984 and 1987, respectively, and the Ph.D. degree in mechanical engineering from the University of Michigan in 1992.

Currently, Dr. Shi is the Carolyn J. Stewart Chair and Professor at the Stewart School of Industrial and Systems Engineering, Georgia Institute of Technology. His research interests include the fusion of advanced statistical and domain knowledge to develop methodologies for modeling, monitoring, diagnosis, and control for complex manufacturing systems.

Dr. Shi is a Fellow of the Institute of Industrial and Systems Engineers (IISE), a Fellow of American Society of Mechanical Engineers (ASME), a Fellow of the Institute for Operations Research and the Management Sciences (INFORMS), an elected member of the International Statistics Institute, a life member of ASA, an Academician of the International Academy for Quality (IAQ), and a member of National Academy of Engineers (NAE).